\documentclass[letterpaper]{article} 
\usepackage{aaai2027}  
\usepackage[hyphens]{url}  
\usepackage{graphicx} 
\usepackage{natbib}  
\usepackage{caption} 
\usepackage{algorithm}
\usepackage{algorithmic}
\usepackage{amsmath}
\usepackage{amssymb}
\usepackage{booktabs}
\usepackage{multirow}
\usepackage{subcaption}
\usepackage{cleveref}
\usepackage{array}

\usepackage{newfloat}
\usepackage{listings}
\DeclareCaptionStyle{ruled}{labelfont=normalfont,labelsep=colon,strut=off} 
\floatstyle{ruled}
\newfloat{listing}{tb}{lst}{}
\floatname{listing}{Listing}

\usepackage{booktabs}

\title{PISA: A Pseudo-Individual Source-Domain Feature Adaptation Framework for Test-Time Open-Vocabulary Object Detection}

\author {
    Ziyan He\textsuperscript{\rm 1},
    Xiongtai Yang\textsuperscript{\rm 1},
    Tao Wang\textsuperscript{\rm 1}\corresponding
}
\affiliations {
    \textsuperscript{\rm 1}Sichuan University\\
    2025223045235@stu.scu.edu.cn, 2025223045170@stu.scu.edu.cn, twangnh@gmail.com
}

\begin{document}

\maketitle

\begin{abstract}
Open-vocabulary object detection test-time adaptation (OVOD-TTA) aims to address the performance degradation that pre-trained base models suffer when encountering image-domain shifts. Existing source-free OVOD-TTA methods rely either on refined test-time information for re-scoring or on pseudo-labels for self-training, leading to significant accuracy degradation when initial predictions are poor. Meanwhile, most conventional source-domain estimation methods recover abstract, sparse representations suitable for the classification task, but fail to capture the dense, concrete features required for detection. To address these issues, we propose \textbf{PISA}, a novel source-free OVOD-TTA method that can be seamlessly integrated into open-vocabulary visual backbones. The core components of our method are the Corruption-Invariant Feature Extractor (CIFE), the Feature Alignment Module (FAM), and a multi-scale alignment framework (BAA). To capture detection-suitable features, we develop CIFE to exploit the invariance of CLIP's visual features across corrupted images, ensuring robustness against various corruptions. We further develop FAM and BAA for the pre-training and adaptation to transform the corruption-invariant features into pseudo-individual source-domain features that are close to the original source-domain features. In this way, dense and concrete pseudo-individual source-domain features are used for supervision instead of unreliable pseudo-label signals. Experiments on the corrupted VOC-C, COCO-C, and LVIS-C benchmarks across three base models demonstrate that PISA substantially improves both the localization precision and the category recognition accuracy of the original models. Notably, PISA achieves state-of-the-art performance without requiring access to source-domain data, surpassing existing methods by 3.92\% in AP@50\% on COCO-C.
\end{abstract}


\section{Introduction}

\begin{figure}[t]
\centering
\includegraphics[width=0.8\columnwidth]{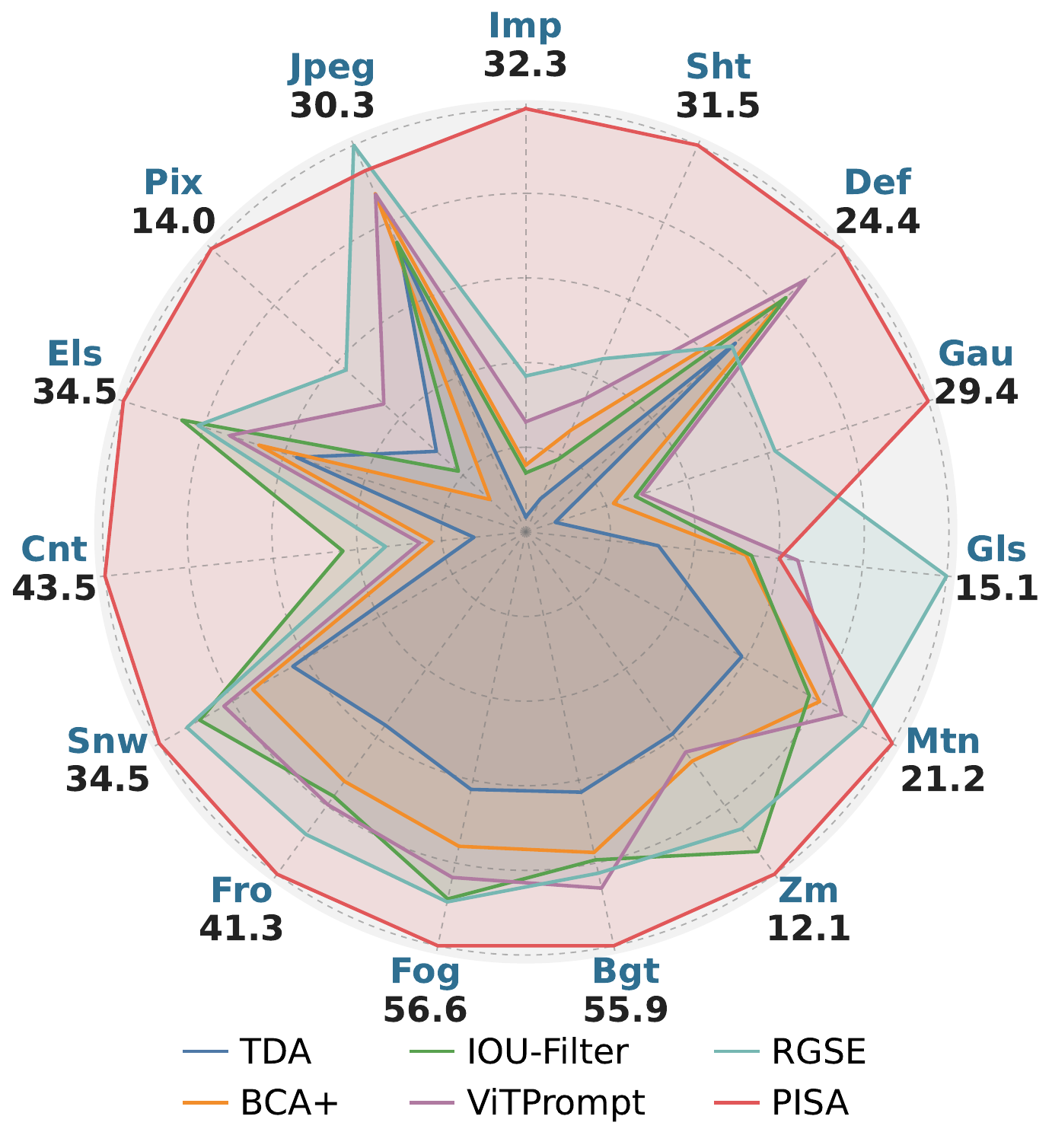} 
\caption{Comparison of PISA with other source free OVOD-TTA methods on COCO-C(AP@50\%) with 15 different kinds of corruptions. PISA achieves state-of-the-art performance across most types of corruptions, demonstrating exceptional robustness.}
\label{fig1}
\end{figure}

Open-vocabulary object detection (OVOD) has emerged as a prominent object detection paradigm \cite{cite1,cite1.1,cite1.2,cite1.3,cite1.4}. By harnessing natural language, OVOD models align text and image features in a shared embedding space via large-scale image-text pre-training, enabling detection of objects beyond a fixed set of categories. However, these models \cite{cite1.5} suffer severe performance degradation when encountering different corrupted images in real scenes. To address this, open-vocabulary object detection test-time adaptation (OVOD-TTA) \cite{cite1.6,cite1.7,cite21} is proposed to improve model's performance by optimizing model during test time. Existing OVOD-TTA methods can be roughly divided into two categories: training-based and training-free methods, as illustrated in Figure \ref{fig:comparison}. Some OVOD-TTA methods \cite{cite27,cite1.9} build upon traditional domain adaptation, aiming to align the distributions of source and target domains during testing, while others rely on pseudo-labels for self-training \cite{cite14,cite17} or cache intermediate information for result aggregation \cite{cite13,cite12,cite35,cite1.6}. Although these methods have their respective advantages, several critical issues remain unresolved:

\begin{figure}[t]
\centering
\includegraphics[width=\columnwidth]{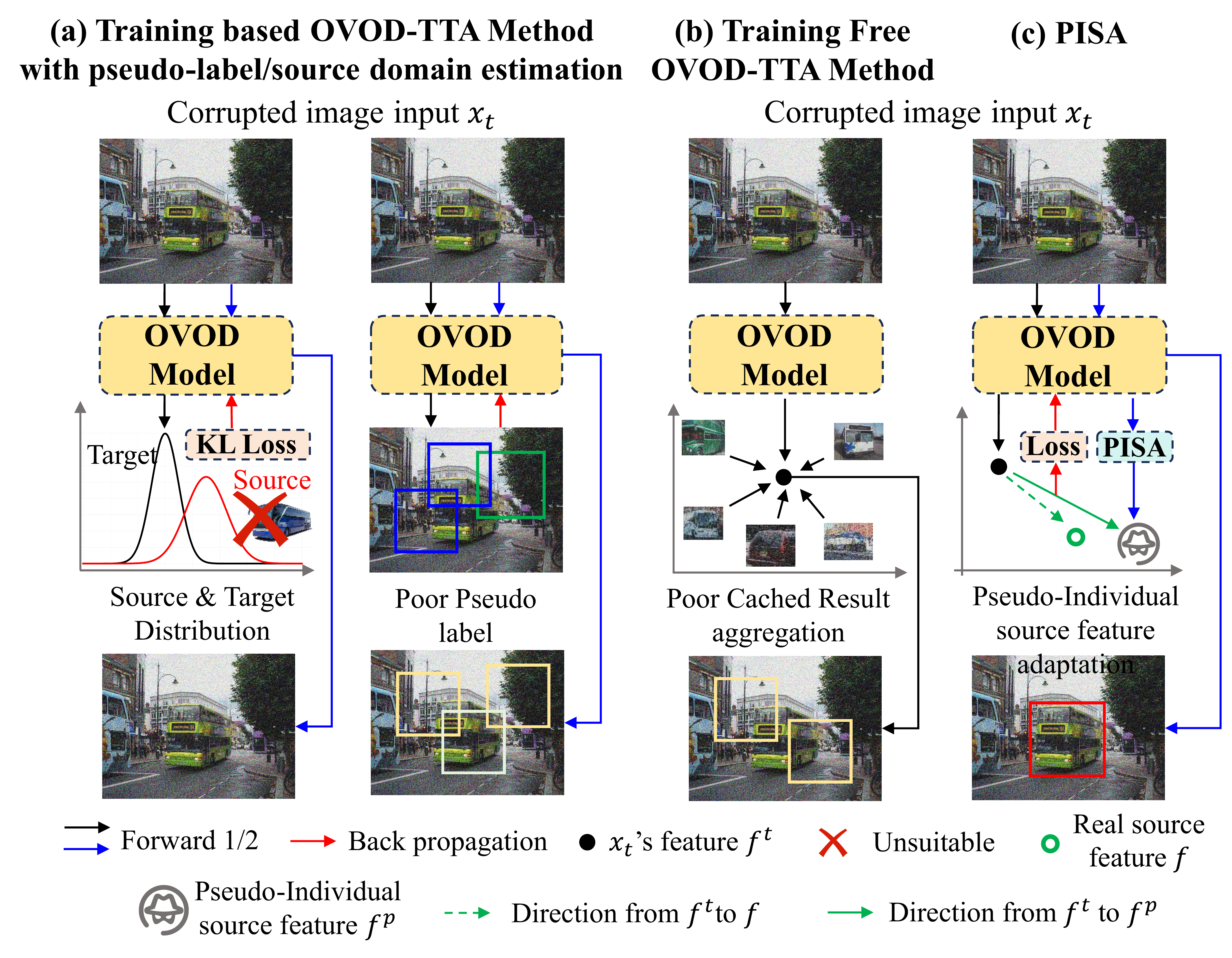} 
\caption{Comparison of different OVOD-TTA methods.  (a) illustrates part of OVOD-TTA method rely on source-domain distribution or pseudo label for self-training. (b) shows the paradigm of cache-based training-free OVOD-TTA mehod.  (c) presents our OVOD-TTA paradigm, which uses pseudo-individual source-domain feature for adaptation.}
\label{fig:comparison}
\end{figure} 

(1)\textbf{How to obtain high-quality, stable supervision signals for model optimization?} For those predictions used as pseudo-labels for self-training or re-scoring the output, severely corrupted test inputs can yield unreliable predictions that progressively degrade the model's capability \cite{cite34}. 

(2)\textbf{How to estimate individual source-domain features without source-domain data?} Obtaining high-quality supervision signals hinges on effective domain estimation. However, conventional source-domain distribution estimation methods are designed for classification and unsuitable for object detection, while accessing source data to estimate distributions violates TTA's core assumption that source features are unavailable at test time \cite{cite1.8}.

To address the above issues, we propose PISA, a novel source-free OVOD-TTA method (Figure 2c). To capture dense, individual features that are suitable and robust for object detection, we leverage the visual backbone of CLIP \cite{cite3}, an open-vocabulary zero-shot classification model, as the Corruption-Invariant Feature Extractor (CIFE). CIFE produces highly similar feature representations for both noisy images and their clean counterparts, which effectively bridges the domain gap. To obtain high-quality signals for model optimization, we develop the Feature Alignment Module (FAM), which consists of a main branch and a target branch. We further develop a multi-scale alignment framework, which mainly consists of two backbone alignment adapters, $\text{BAA}_{t}$ and $\text{BAA}_{a}$. $\text{BAA}_{t}$ transfers corruption-invariant features into pseudo instance-level source-domain features, while $\text{BAA}_{a}$ is the only component kept trainable during test time, with all other components frozen to ensure stable model updates. In this manner, PISA attains stable optimization of the model at test time through the use of pseudo-individual source-domain features. As shown in Figure \ref{fig1}, PISA performs the best among existing source free OVOD-TTA methods on COCO-C with 15 different kinds of corruptions. More importantly, PISA is a general framework that can be applied to many OVOD backbones. Our main contributions are summarized as follows:

\begin{itemize}
    \item We propose PISA, a source-free OVOD-TTA framework based on pseudo-individual source-domain features, which substantially improves the original OVOD model's performance across diverse corruptions. 
    \item We employ CLIP in OVOD-TTA and propose the Corruption-Invariant Feature Extractor (CIFE) for individual source-domain feature estimation, which extracts the corruption-invariant feature suitable for object detection from different corrupted images and bridges the domain gap. 
    \item We propose Feature Alignment Module (FAM) and a multi-scale alignment framework (BAA), transferring CIFE features into high-quality pseudo-individual source-domain features for stable model optimization. 
    \item We evaluate PISA with three OVOD methods on VOC-C, COCO-C, and LVIS-C. The results demonstrate that PISA substantially improves the performance of multiple baseline models. Specifically, PISA outperforms the state of the art by 3.92\% on COCO-C.
\end{itemize}

\begin{figure*}[ht]
\centering 
\includegraphics[width=\textwidth]{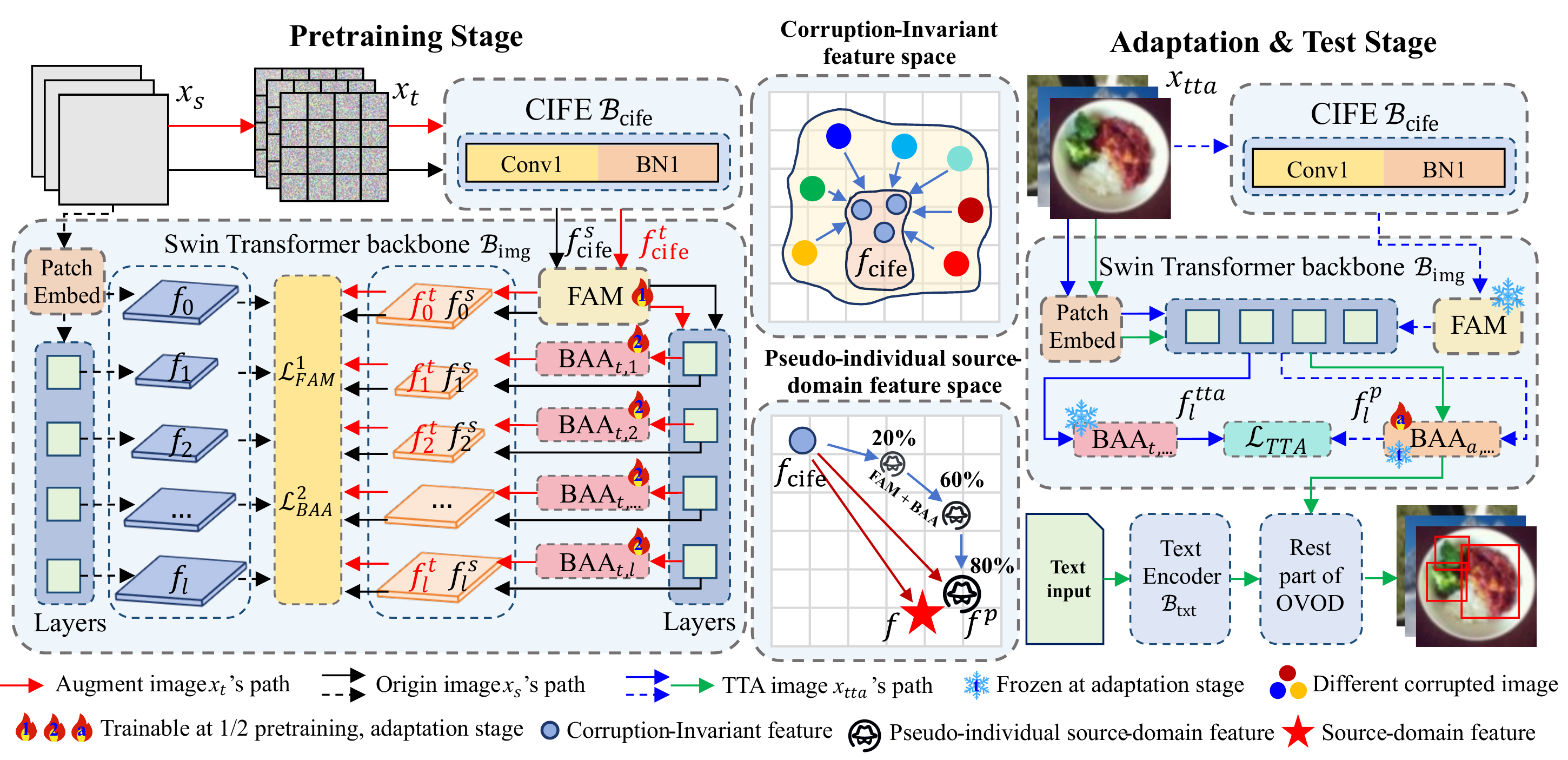} 
\caption{Overview of PISA. The CIFE module extracts corruption-invariant features, while the FAM aligns them with patch embeddings via main and target branches. BAA integrates multi-scale adapters into the backbone to bridge the domain gap.} 
\label{fig:picture002} 
\end{figure*} 

\section{Related Work}
\subsection{Source-Domain Estimation}
Several early works explore reconstructing a surrogate source feature distribution without source data. \citet{SF1} estimate the source distribution via statistical descriptors from the pre-trained model and align target features toward it. VDM-DA \cite{SF2} synthesizes pseudo source features using prototype information encoded in the source model. Similarly, \citet{SF3} build a pseudo-source domain through generation and augmentation of synthetic features from the source model, enabling source-free adaptation. However, they are predominantly designed for image-level classification, where only abstract, global representations are needed. In contrast, detection models rely on dense, concrete features, making it more challenging to reconstruct the source domain. To address this, our method reconstructs multi-scale pseudo-individual source-domain features, enabling better adaptation to the dense feature requirements of object detection.

\subsection{Source-Free OVOD-TTA}
OVOD-TTA refers to adapting the object detection model during the testing phase. Source free OVOD-TTA methods eliminate the need for source domain data, performing adaptation exclusively on target domain data at test time. Existing approaches can be categorized into training-free and training-based methods. The former mainly rely on cache-driven frameworks, which store refined visual features, text embeddings, or other test-time information to re-score the initial predictions, as in TDA \cite{cite13}, BCA \cite{cite12}, BCA+\cite{cite11}, ViTPrompt \cite{cite17}, RGSE\cite{cite35}, FACTOR\cite{cite1.6}. The latter perform incremental self-supervised fine-tuning on test samples by updating only a small subset of parameters, establishing them as the mainstream paradigm for adaptive detection. IOU-Filter \cite{cite14} is the first fully self-supervised OVOD-TTA framework, which refines training samples by discarding low-confidence pseudo-labels via thresholding. VLOD-TTA \cite{cite2.1} strengthens domain-shift robustness through IoU-weighted entropy minimization on spatially coherent proposals, coupled with image-conditioned prompt selection.  However, both methods are sensitive to output quality, becoming less effective in complex environments when pseudo-labels or cached features are unreliable. In contrast, our approach achieves source-free TTA while leveraging target-domain data to ensure high-quality optimization. 

\section{Methodology}

\subsection{Preliminary}
Since open-vocabulary object detectors already exhibit strong capability in recognizing novel categories, we primarily focus on the TTA scenario under different corruptions. Suppose we have a pre-trained OVOD model on the source-domain data, which consists of a text encoder $\mathcal{B}_\text{txt}$ and an image encoder $\mathcal{B}_\text{img}$. The key task is to fine-tune the model using unlabeled target-domain data $\{x_i^{tta}\}_{i=1}^{N_t}$ to generalize it to better predict in target domain \cite{cite21}. In this section, we first review the traditional TTA paradigm and present the overall framework of our method. We then elaborate on the design of each module and describe the detailed procedures for every stages.

\subsubsection{Traditional OD-TTA Method}
 Traditional OD-TTA methods, based on a self-supervised learning paradigm, rely on pseudo-labels for online optimization. When test sample $x_{tta}$ arrives, the original model $\mathcal{B}_\theta$ first generates the  prediction set $\{b_t^i,y_t^i\}_{i=0}^N= \mathcal{B}_\theta(x_{tta})$. High-confidence predictions are then utilized as pseudo-labels to minimize a self-training loss, thereby updating the model parameters from $\theta$ to $\theta^{*}$:

\begin{equation}
    \hat{y}_t=\mathbb{I}\left( p_t>\tau \right) \mathcal{B}_{\theta }\left( x_{tta} \right) 
\end{equation}

\begin{equation}
\theta ^*=\text{arg}\min_{\theta}\sum_{x_t\in D_t}{\mathcal{L}\left( \hat{y}_t \right)}
\end{equation}
Although such a paradigm can mitigate the distribution shift between domains, its performance remains critically dependent on pseudo-label quality. On images with substantial domain shifts, erroneous pseudo-labels readily accumulate throughout iterative updates, causing the model to assimilate these spurious patterns.

\subsection{Proposed Method}
To avoid the issue of low-quality pseudo-labels in the test domain, we instead resort to feature-level supervision, where the model is optimized to generate source-like features with a pseudo-feature alignment objective. 
The complete pipeline of our method is illustrated in Figure \ref{fig:picture002}, which mainly comprises three components: the Corruption-Invariant Feature Extractor (CIFE), the Feature Alignment Module (FAM) and a multi-scale alignment framework ($\text{BAA}_t$ and $\text{BAA}_a$). The CIFE is responsible for extracting corruption-invariant features $f_\text{cife}$, while the FAM and $\text{BAA}_t$ modules are designed to transform $f_\text{cife}$ into pseudo-individual source-domain features $f^{p}$ that is close enough to the real source-domain features $f$. As shown in Figure \ref{fig:picture002}, within the pseudo-individual source-domain feature space, after alignment via FAM and BAA, the direction from $f_\text{cife}$ to $f^{p}$ closely aligns with that from $f_\text{cife}$ to $f$. Thus, $\text{BAA}_a$ is used to fine-tune the original backbone with $f^{p}$ during testing. Here, $\mathcal{B}_{\text{img},\theta}$ denotes the visual backbones of the OVOD model, and $F_{\text{pisa}}$ represents our method.

\begin{equation}
    \theta ^*=\text{arg}\min_{\theta}\sum_{x_t\in D_t}{\mathcal{L}\left( \mathcal{B}_{\text{img},\theta}\left( x_{tta} \right) ,\mathcal{B}_{\text{img},\theta}\left( F_{\text{pisa}}\left( x_{tta} \right) \right) \right)}
\end{equation}

\begin{figure}
    \centering
    \includegraphics[width=\linewidth]{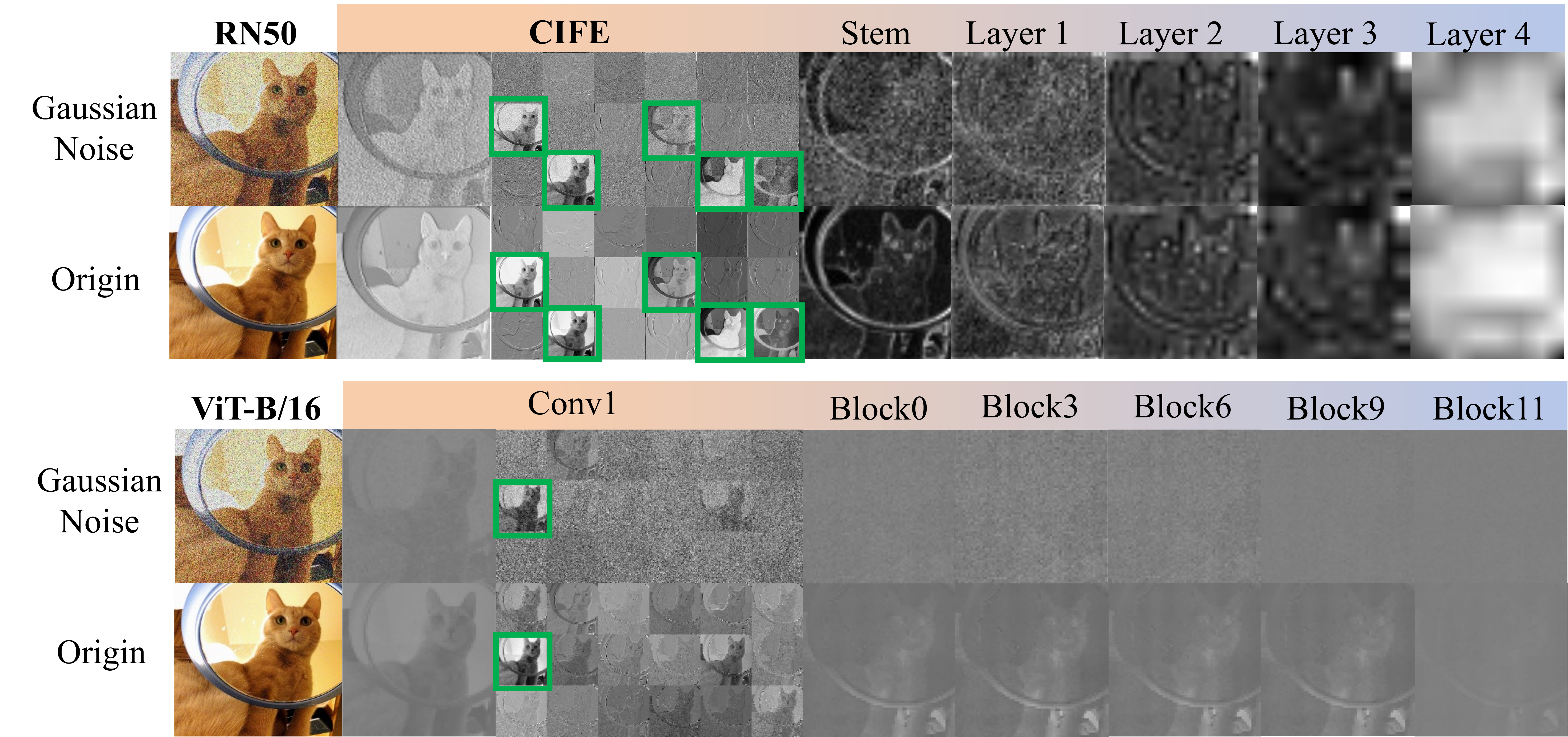}
    \caption{Visualization of the outputs from CLIP visual backbones: ResNet-50 (top) and ViT-B/16 (bottom). Green bboxes highlight features that similar to original images'.}
    \label{visual}
\end{figure}
After fine-tuning the model with the input image $x_t$, the updated parameters $\theta^*$ are applied to re-evaluate the current image $x_{tta}$ to obtain the final predictions $\mathcal{B}_{\text{img},\theta^*}(x_{tta})$.

\subsection{Corruption-Invariant Feature Extractor(CIFE)}
Trained on massive data, CLIP exhibits inherent robustness to corruption. For instance, ResNet-series CLIP backbones preserve feature representations that remain highly similar to those extracted from clean images, even under varying levels of Gaussian noise \cite{cite22}. Furthermore, we conducted visual analyses on images sharing identical content but subjected to different noise perturbations. As shown in Figure \ref{visual}, from left to right, features transition from dense and concrete to abstract and sparse. Shallow features retain both noise robustness and original image information, whereas deep features, despite stronger robustness, lose much of the visual detail. Moreover, compared to ViT-B/16, features extracted by ResNet-50 demonstrate superior robustness against noise, as evidenced by the green boxes. Motivated by this observation, to preserve more raw pixel-level information, we construct a Corruption-Invariant Feature Extractor (CIFE), which consists of the first convolution and batch normalization layer of the stem structure of CLIP's RN50 visual backbone. CIFE can extract features that are invariant to object style and noise while preserving most of the original image information, which are termed corruption-invariant features. As illustrated in Figure \ref{fig:picture002}, within the corruption-invariant feature space, variations in corruption-invariant features across differently corrupted images are negligible. Additionally, we do not apply normalization to the input images, as this would amplify the impact of corruption.

\begin{table*}[t]
\centering
\small
\setlength{\tabcolsep}{2.2pt} 
\begin{tabular*}{\textwidth}{@{\extracolsep{\fill}} c | c | cc @{\hspace{10pt}} cc @{\hspace{10pt}} cc}
\toprule
\multirow{2}{*}{Dataset} & \multirow{2}{*}{Method} & \multicolumn{2}{c}{Base} & \multicolumn{2}{c}{Novel} & \multicolumn{2}{c}{All} \\
\cmidrule(lr){3-4} \cmidrule(lr){5-6} \cmidrule(lr){7-8}
 & & AP@50\% & mAP & AP@50\% & mAP & AP@50\% & mAP \\
\midrule
\multirow{6}{*}{VOC}
 & GLIP \cite{cite4}          & 17.86 & 12.40 & 36.17 & 25.40 & 33.68 & 23.63 \\
 & PISA (Ours)     & \textbf{35.98} \textcolor[RGB]{0,100,0}{\scriptsize(+18.12)} & \textbf{24.25} \textcolor[RGB]{0,100,0}{\scriptsize(+11.85)} & \textbf{43.31} \textcolor[RGB]{0,100,0}{\scriptsize(+7.14)} & \textbf{30.24} \textcolor[RGB]{0,100,0}{\scriptsize(+4.84)} & \textbf{42.31} \textcolor[RGB]{0,100,0}{\scriptsize(+8.63)} & \textbf{29.42} \textcolor[RGB]{0,100,0}{\scriptsize(+5.79)} \\
 & GDINO \cite{cite5}         & 31.00 & 22.60 & 44.77 & 33.86 & 42.90 & 32.33 \\
 & PISA (Ours)    & \textbf{48.33} \textcolor[RGB]{0,100,0}{\scriptsize(+17.33)} & \textbf{35.17} \textcolor[RGB]{0,100,0}{\scriptsize(+12.57)} & \textbf{48.61} \textcolor[RGB]{0,100,0}{\scriptsize(+3.84)} & \textbf{36.63} \textcolor[RGB]{0,100,0}{\scriptsize(+2.77)} & \textbf{48.57} \textcolor[RGB]{0,100,0}{\scriptsize(+5.67)} & \textbf{36.42} \textcolor[RGB]{0,100,0}{\scriptsize(+4.09)} \\
 & LLMDet \cite{cite7}        & 31.77 & 21.30 & 50.44 & 36.46 & 47.90 & 34.39 \\
 & PISA (Ours)   & \textbf{49.67} \textcolor[RGB]{0,100,0}{\scriptsize(+17.90)} & \textbf{34.13} \textcolor[RGB]{0,100,0}{\scriptsize(+12.83)} & \textbf{53.98} \textcolor[RGB]{0,100,0}{\scriptsize(+3.54)} & \textbf{39.12} \textcolor[RGB]{0,100,0}{\scriptsize(+2.66)} & \textbf{53.40} \textcolor[RGB]{0,100,0}{\scriptsize(+5.50)} & \textbf{38.44} \textcolor[RGB]{0,100,0}{\scriptsize(+4.05)} \\
\midrule
\multirow{6}{*}{COCO}
 & GLIP \cite{cite4}         & 9.83  & 6.34  & 21.58 & 14.25 & 19.97 & 13.18 \\
 & PISA (Ours)     & \textbf{23.90} \textcolor[RGB]{0,100,0}{\scriptsize(+14.07)} & \textbf{15.67} \textcolor[RGB]{0,100,0}{\scriptsize(+9.33)} & \textbf{25.25} \textcolor[RGB]{0,100,0}{\scriptsize(+3.67)} & \textbf{16.70} \textcolor[RGB]{0,100,0}{\scriptsize(+2.45)} & \textbf{25.07} \textcolor[RGB]{0,100,0}{\scriptsize(+5.10)} & \textbf{16.56} \textcolor[RGB]{0,100,0}{\scriptsize(+3.38)} \\
 & GDINO \cite{cite5}        & 22.53 & 15.70 & 30.39 & 21.21 & 29.32 & 20.45 \\
 & PISA (Ours)    & \textbf{32.50} \textcolor[RGB]{0,100,0}{\scriptsize(+9.97)} & \textbf{22.37} \textcolor[RGB]{0,100,0}{\scriptsize(+6.67)} & \textbf{33.30} \textcolor[RGB]{0,100,0}{\scriptsize(+2.91)} & \textbf{23.17} \textcolor[RGB]{0,100,0}{\scriptsize(+1.96)} & \textbf{33.20} \textcolor[RGB]{0,100,0}{\scriptsize(+3.88)} & \textbf{23.06} \textcolor[RGB]{0,100,0}{\scriptsize(+2.61)} \\
 & LLMDet \cite{cite7}       & 23.07 & 16.13 & 32.31 & 22.71 & 31.05 & 21.81 \\
 & PISA (Ours)   & \textbf{33.20} \textcolor[RGB]{0,100,0}{\scriptsize(+10.13)} & \textbf{23.07} \textcolor[RGB]{0,100,0}{\scriptsize(+6.94)} & \textbf{34.67} \textcolor[RGB]{0,100,0}{\scriptsize(+2.36)} & \textbf{24.33} \textcolor[RGB]{0,100,0}{\scriptsize(+1.62)} & \textbf{34.47} \textcolor[RGB]{0,100,0}{\scriptsize(+3.42)} & \textbf{24.16} \textcolor[RGB]{0,100,0}{\scriptsize(+2.35)} \\
\midrule
\multirow{4}{*}{LVIS}
 & GLIP \cite{cite4}         & 5.40  & 3.33  & 10.62 & 7.01  & 9.90  & 6.51  \\
 & PISA (Ours)    & \textbf{10.23} \textcolor[RGB]{0,100,0}{\scriptsize(+4.83)} & \textbf{6.63} \textcolor[RGB]{0,100,0}{\scriptsize(+3.30)} & \textbf{11.64} \textcolor[RGB]{0,100,0}{\scriptsize(+1.02)} & \textbf{7.70} \textcolor[RGB]{0,100,0}{\scriptsize(+0.69)} & \textbf{11.45} \textcolor[RGB]{0,100,0}{\scriptsize(+1.55)} & \textbf{7.55} \textcolor[RGB]{0,100,0}{\scriptsize(+1.04)} \\
 & GDINO \cite{cite5}        & 8.93  & 5.80  & 12.89 & 8.75  & 12.35 & 8.35  \\
 & PISA (Ours)    & \textbf{13.10} \textcolor[RGB]{0,100,0}{\scriptsize(+4.17)} & \textbf{8.67} \textcolor[RGB]{0,100,0}{\scriptsize(+2.87)} & \textbf{14.17} \textcolor[RGB]{0,100,0}{\scriptsize(+1.28)} & \textbf{9.66} \textcolor[RGB]{0,100,0}{\scriptsize(+0.91)} & \textbf{14.03} \textcolor[RGB]{0,100,0}{\scriptsize(+1.68)} & \textbf{9.52} \textcolor[RGB]{0,100,0}{\scriptsize(+1.17)} \\
\bottomrule
\end{tabular*}
\caption{Object detection performance compared with baseline models on VOC-C, COCO-C, and LVIS-C.}
\label{tab:table1}
\end{table*}

\subsection{Feature Alignment Module (FAM)}
\begin{figure}[t]
    \centering
    \includegraphics[width=0.9\linewidth]{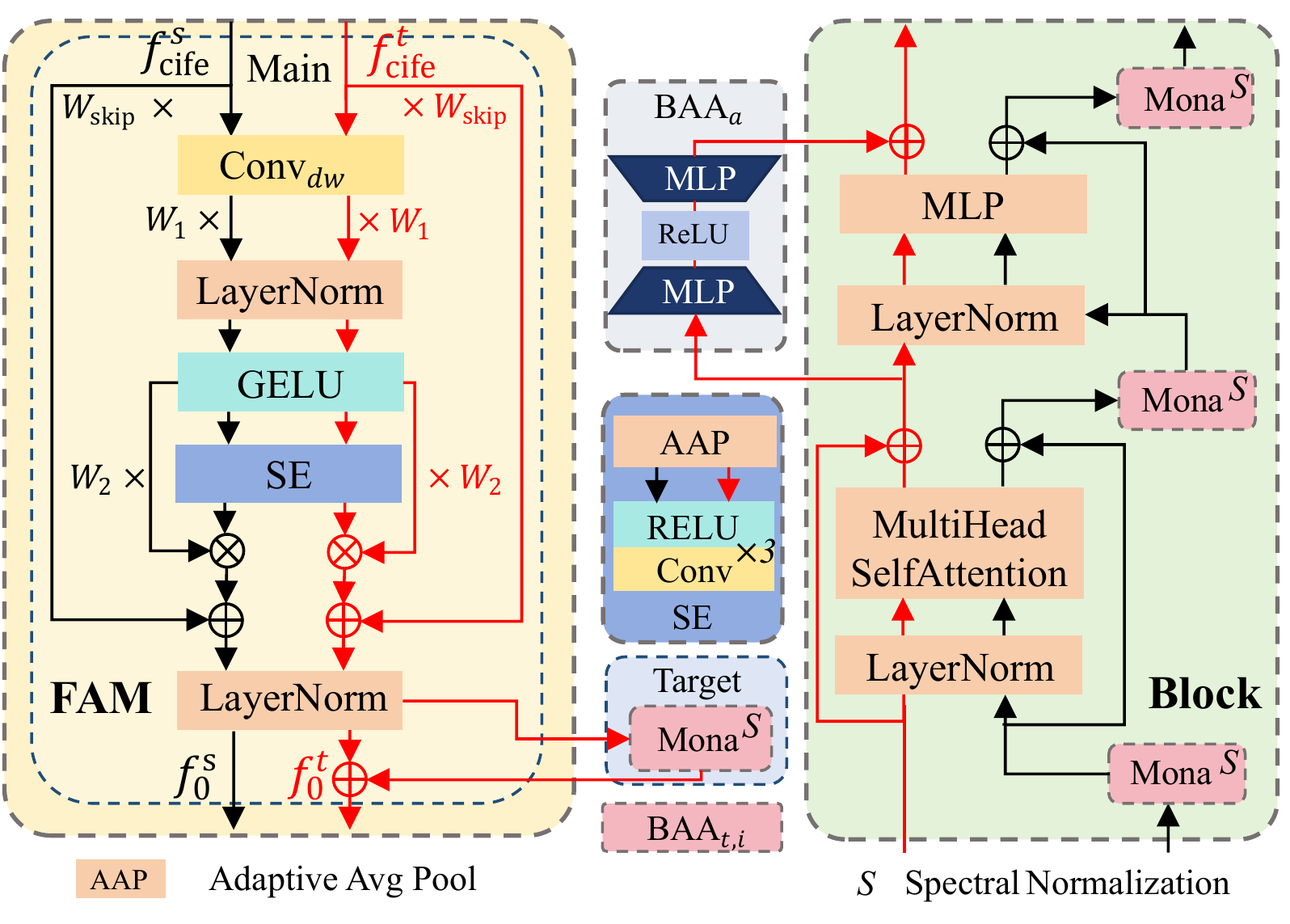}
    \caption{The architecture of FAM and the inserted position of $\text{BAA}_t$ (last block) and $\text{BAA}_a$ (every block) modules.}
    \label{FAM}
\end{figure}

Given the source and target CIFE features $f_{\mathrm{cife}}$, FAM aligns the inputs with the patch embedding output while maintaining robustness to corrupted target samples. As shown in Figure \ref{FAM}, it consists of a main branch and a target branch.

\paragraph{Main Branch.}
The main branch jointly performs spatial downsampling and channel
projection.  We first apply a depth-wise separable convolution to
reduce parameters while preserving spatial locality:
\begin{equation}
  \tilde{f}
  = \mathrm{GELU}\!\Big(\mathrm{LN}\big(
      W_1\cdot \mathrm{Conv}_{dw}(f_{\mathrm{cife}})
    \big)\Big),
  \label{eq:main_down}
\end{equation}
followed by a Squeeze-and-Excitation (SE) attention that recalibrates channel-wise responses via a bottleneck $\mathcal{F}_{\mathrm{SE}}$. SE attention suppresses domain-specific noise channels and enhances transferable ones. A residual shortcut then fuses the projected feature with the input, where $W_{\mathrm{skip}}$ ensures dimensional alignment.  The shortcut guarantees stable gradient flow during the pre-training stage.
\begin{equation}
  \hat{f} = \tilde{f} \odot \mathcal{F}_{\mathrm{SE}}(\tilde{f}),  f_0 = \mathrm{LN}\!\big(W_{\mathrm{2}}\,\hat{f} + W_{\mathrm{skip}}\,f_{\mathrm{cife}}\big)
  \label{eq:se}
\end{equation}

\paragraph{Target Branch.}
To further handle corrupted samples, we optionally pass $f_{0}^{t}$ through Mona \cite{cite26}, a multi-scale visual adapter. To enhance the network's robustness against input perturbations, we apply spectral normalization to each convolutional layer within Mona by re-parameterizing its weight tensor $W_m$ as:
\begin{equation}
  \hat{W}_m = \frac{W_m}{\sigma(W_m)},
  \quad
  \sigma(W_m) = \max_{\|h\|_2 \leq 1} \|W_m\, h\|_2,
  \label{eq:spec_norm}
\end{equation}
where $\sigma(W_m)$ is the largest singular value of $W_m$.

\subsection{Multi-scale Alignment Framework}
\subsubsection{Backbone Alignment Adapter (BAA)}
Unfortunately, the Swin Transformer backbone of $\mathcal{B}_\text{img}$ tends to amplify the discrepancies between similar input samples, thereby progressively exacerbating the gap between corruption-invariant features and source-domain features. To address this issue, we follow \cite{cite36} and construct a series of multi-scale adapters $\text{BAA}_{t}$. Each consists of two components: (1) the same Mona \cite{cite26} adapter as in FAM, inserted at the last block of each layer; and (2) the module that adjusts the dimensionality of the backbone features at each layer. As pseudo-individual source-domain features offer only directions roughly aligned with the source domain, we insert the TTA adapter $\text{BAA}_{a}$ \cite{cite27} into each block for better test-time adaptation.

\subsubsection{Pre-training Stage}
During pre-training, our method relies solely on clean and augmented image pairs $( x_s, x_t)$. $x_s$ is fed into the visual backbone of the open-vocabulary object detector to obtain the source-domain feature set $\{{f_l}\}_{l=1}^L$. Meanwhile, both $x_s$ and $x_t$ are processed by $\mathcal{B}_\text{cife}$, FAM, and the remaining parts of $\mathcal{B}_\text{img}$ (excluding the patch embedding) to yield $\{{f_l^s}\}_{l=1}^L$ and $\{{f_l^t}\}_{l=1}^L$, respectively. We conduct pre-training in two stages. In the first stage, we train the FAM to align the outputs of $\mathcal{B}_\text{cife}$+FAM+$\mathcal{B}_\text{img}$ (without passing $\text{BAA}_t$ in $\mathcal{B}_\text{img}$). In the second stage, we freeze the trained FAM and solely train $\text{BAA}_t$ ($x_t$ additionally passes through $\text{BAA}_t$ in $\mathcal{B}_\text{img}$). The two stages utilize the loss functions $\mathcal{L}_{FAM}^1$ and $\mathcal{L}_{BAA}^2$ for alignment (where $\alpha$ is a hyperparameter used to prevent excessive feature discrepancies), respectively:
\begin{equation}
    \mathcal{L}_{FAM}^1 = \sum_{l=0}^L \left( |f_l^t - f_l| + |f_l^s - f_l| + \alpha |f_l^s - f_l^t| \right) 
\end{equation}
\begin{equation}
    \mathcal{L}_{BAA}^2 = \sum_{l=1}^L |f_l^t - f_l|
\end{equation}

\begin{table*}[ht]
\centering
\small
\setlength{\tabcolsep}{2.2pt}
\begin{tabular*}{\textwidth}{@{\extracolsep{\fill}} c | c | ccccccccccccccc c @{}}
\toprule
Method & Source & Imp & Sht & Gau & Def & Gls & Mtn & Zm & Bgt & Fog & Fro & Snw & Cnt & Els & Pix & Jpeg & AVG \\
\midrule
TDA             & CVPR2024  & 16.7 & 17.1 & 15.8 & 20.3 & 9.9  & 16.9 & 9.6  & 45.5 & 45.9 & 32.3 & 28.2 & 24.4 & 27.1 & 9.0  & 26.3 & 23.00 \\
HisTPT           & NeuralPS2024  & 18.2 & 17.9 & 17.6 & 21.3 & \textbf{\textcolor{gray!80}{12.5}} & 19.1 & 9.9  & 45.6 & 45.1 & 32.8 & 28.2 & 26.1 & 26.0 & 8.6  & 26.2 & 23.67 \\
IOU-Filter      & CVPR2024 & 18.4 & 18.7 & 18.7 & 22.3 & 11.6 & 18.8 & \textbf{\textcolor{gray!80}{11.7}} & 50.1 & \textbf{\textcolor{gray!80}{53.4}} & 36.6 & 32.6 & \textbf{\textcolor{gray!80}{31.2}} & \textbf{\textcolor{gray!80}{32.0}} & 8.5  & 26.5 & 26.07 \\
BCA             & CVPR2025 & 17.3 & 18.0 & 16.0 & 21.0 & 10.9 & 16.1 & 9.9  & 46.3 & 46.7 & 33.9 & 29.0 & 25.2 & 27.5 & 8.9  & 26.2 & 23.53 \\
BCA+            & ARXIV2025 & 18.7 & 19.9 & 17.9 & 22.2 & 11.5 & 19.1 & 10.1 & 49.6 & 49.8 & 35.7 & 30.1 & 26.6 & 28.7 & 7.8  & 28.4 & 25.07 \\
ViTPrompt      & CVPR2026 & 20.4 & 21.2 & 18.9 & 23.1 & 12.4 & 19.7 & 9.9  & \textbf{\textcolor{gray!80}{52.0}} & 51.9 & 37.1 & 31.5 & 27.2 & 30.0 & 10.2 & 28.4 & 26.26 \\
RGSE            & ARXIV2026 & \textbf{\textcolor{gray!80}{22.1}} & \textbf{\textcolor{gray!80}{22.8}} & \textbf{\textcolor{gray!80}{20.2}} & \textbf{\textcolor{gray!80}{23.8}} & \textbf{15.1} & \textbf{\textcolor{gray!80}{20.3}} & 11.3 & 51.0 & 53.6 & \textbf{\textcolor{gray!80}{38.9}} & \textbf{\textcolor{gray!80}{33.2}} & 29.0 & 31.3 & \textbf{\textcolor{gray!80}{11.0}} & \textbf{{30.3}} & 27.59 \\
PISA                         & Ours     & \textbf{32.3} & \textbf{31.5} & \textbf{29.4} & \textbf{24.4} & 12.1 & \textbf{21.2} & \textbf{12.1} & \textbf{55.9} & \textbf{56.6} & \textbf{41.3} & \textbf{34.5} & \textbf{43.5} & \textbf{34.5} & \textbf{14.0} & \textbf{\textcolor{gray!80}{29.3}} & \textbf{31.51}\textcolor[RGB]{0,100,0}{\scriptsize(+3.92)} \\
\bottomrule
\end{tabular*}
\caption{Object Detection Performance Comparison with State-of-the-Art Methods on COCO-C. Metric: mean average precision at 50\% (AP@50\%). The best results are denoted in bold and the second best results are denoted in bold gray.}
\label{tab:table2}
\end{table*}

\begin{figure*}
    \centering
    \includegraphics[width=\textwidth]{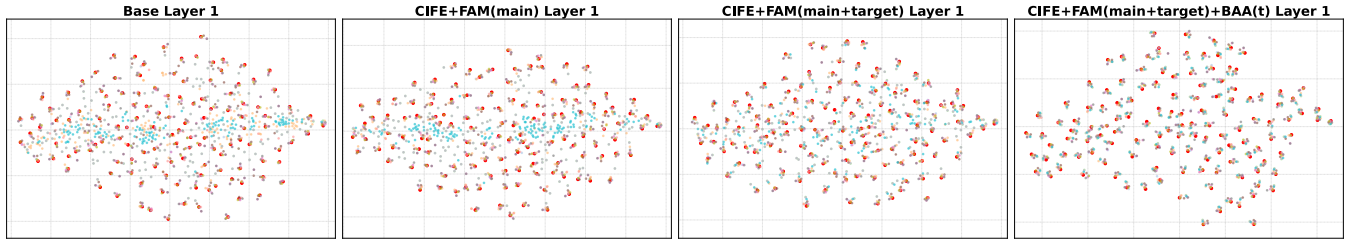}
    \caption{t-SNE visualization of clean (red) and corrupted (other colors) feature distributions at the first output layer of different PISA modules. The distinct cluster-like distribution in the right plot indicates that pseudo-individual source-domain features closely resemble real source-domain features.}
    \label{tsne_ablation}
\end{figure*}

\subsubsection{Adaptation Stage}
After the pre-training stage, for a test-time input $x_{tta}$, we process it through $\mathcal{B}_\text{cife} + \text{FAM} + \mathcal{B}_\text{img}$ (excluding the patch embedding and utilizing $\text{BAA}_{t}$) and through $\mathcal{B}_\text{img}$ (utilizing $\text{BAA}_{a}$) to obtain pseudo-individual source-domain feature $\{f_l^{p}\}_{l=1}^L$ and $x_{tta}$'s feature $\{f_l^{tta}\}_{l=1}^L$, respectively. Finally, we employ a cosine similarity loss function to update the $\text{BAA}_{a}$ module:
\begin{equation}
    \mathcal{L}_{TTA} = \sum_{l=1}^L \left( 1 - \frac{f_{l}^p \cdot f_l^{tta}}{\|f_l^{p}\|_2 \|f_l^{tta}\|_2} \right)
\end{equation}
After updating the trainable parameters ($\text{BAA}_a$), we evaluate $x_{tta}$ on the model with the updated $\text{BAA}_a$ to obtain the final predictions. Furthermore, to prevent catastrophic forgetting, we reset the model to its initial state every $T$ test samples.

\section{Experiment}
We aim to investigate the following questions in experiment part: (1) Does PISA exhibit strong generalizability? (2) How does PISA compare with state-of-the-art approaches? and (3) How do hyper parameter affect the performance of PISA?

\subsection{Experimental Setup}
\subsubsection{Datasets} We utilize three datasets with different category sizes: VOC-C(20), COCO-C(80), and LVIS-C(1203). Based on COCO-2017 and VOC2007, and following \cite{cite29,cite37,cite30,cite31}, we artificially introduce 19 types of corruptions and 3 types of styles to construct the corrupted datasets. Furthermore, since LVIS-minival \cite{cite33} and COCO share the same images, we also conduct evaluations on the LVIS-minival annotations using the aforementioned COCO-C datasets.

\subsubsection{Implementation Details} During the pre-training stage, we selected 10,000 unlabeled images from the COCO-2017 dataset \cite{cite28}, 10,000 images from the training and validation sets of VOC2012 \cite{cite32}, and 5,011 images from the training and validation sets of VOC2007\cite{cite32}, totaling 25,011 images. We chose the most commonly used data robustness operations: Gaussian noise, Gaussian blur, and contrast—as the base types to augment the training data (for each image, one of the three corruptions and one of the five severity levels from 1 to 5 are randomly selected). The remaining 19 are referred to as novel types. Both FAM and BAA are trained using AdamW for 10 epochs with a batch size of 4 and the pre-training learning rate ${lr}_p$ of 0.01. $\alpha$ is set to 10. For noise and contrast corruption, the adaptation learning rate ${lr}_a$ is set to 0.01. For saturation, motion blur, oil painting, pencil color and pencil grey, ${lr}_a$ is set to 0.005. For the rest corruptions, ${lr}_a$ is set to 0.001. We set $T$ to $\{1, 2, 4, 8, 16, \infty\}$ and report the best result for each corruption. All pre-training and adaptation experiments are conducted on a single NVIDIA RTX 4090 GPU.

\subsection{Comparison Result}
\subsubsection{Baseline Model Comparison}
We evaluate the baseline methods by integrating PISA with an open-vocabulary visual backbone. As shown in Table \ref{tab:table1}, our method outperforms all baseline methods across the three corrupted datasets, which exhibit strong generalizability across different models and datasets. Across the three models, PISA improves the average mAP on base corruption types by 12.42\% on VOC-C, 7.65\% on COCO-C, and 3.09\% on LVIS-C. Despite being pre-trained on only three corruption types, it also demonstrates strong generalization to unseen corruption types, with average mAP gains of 3.42\% (VOC-C) and 2.01\% (COCO-C). Overall, the average mAP improvements across all corruption types are 4.64\%, 2.78\%, and 1.11\% on VOC-C, COCO-C, and LVIS-C, respectively. These results demonstrate that, despite the increasing number of object categories, PISA still yields substantial performance gains over the original methods.

\begin{figure*}[ht]
\centering 
\includegraphics[width=\textwidth]{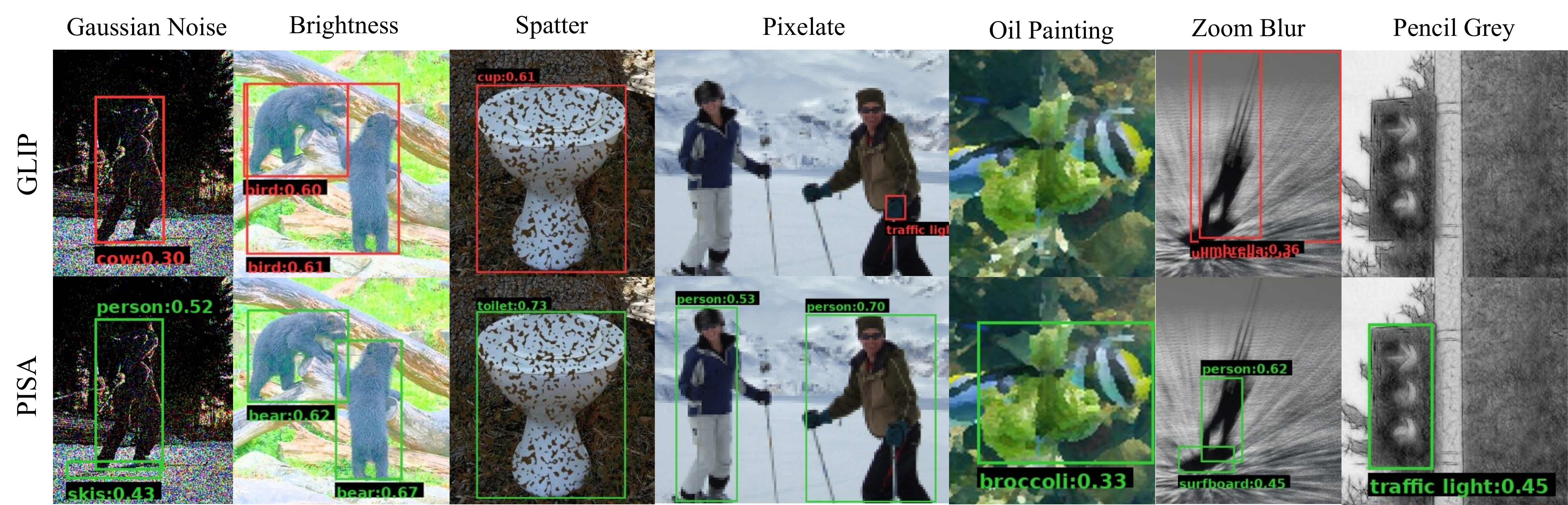} 
\caption{Qualitative comparison against the baseline GLIP. Bounding boxes are overlaid on the images. Red and green boxes indicate negative and positive results, respectively, with an IoU threshold of 0.5.} 
\label{fig:5} 
\end{figure*} 

\subsubsection{State-of-the-art Comparison}
The robust object detection results in Table \ref{tab:table2} show that PISA consistently outperforms state-of-the-art methods across most corruption types on COCO-C datasets. It achieves the highest average AP@50\% of 31.51\%, surpassing the best baseline RGSE by 3.92\%. Notably, PISA achieves clear gains on both severe noise and digital corruptions, indicating strong robustness to diverse perturbations. These results highlight PISA's exceptional robustness and stability in handling diverse and severe real-world image degradations.

\begin{table}[ht]
\centering
\footnotesize
\setlength{\tabcolsep}{2pt}
\begin{tabular}{ccccc|ccc}
\toprule
\multirow{2}{*}{Base} & \multicolumn{2}{c}{FAM} & \multirow{2}{*}{$\text{BAA}_t$} & \multirow{2}{*}{$\text{BAA}_a$} & \multirow{2}{*}{AP@50\%} & \multirow{2}{*}{AP@75\%} & \multirow{2}{*}{mAP} \\
 \cmidrule(lr){2-3} & Main & Target & & & & & \\
\midrule
$\checkmark$ & & & & & 29.32 & 21.99 & 20.45 \\
$\checkmark$ & $\checkmark$ & & & $\checkmark$ & 31.46 & 23.66 & 22.00 \\
$\checkmark$ & $\checkmark$ & $\checkmark$ & & & 31.54 & 23.25 & 21.74 \\
$\checkmark$ & $\checkmark$ & $\checkmark$ & & $\checkmark$ & 32.84 & 24.62 & 22.92 \\
$\checkmark$ & $\checkmark$ & & $\checkmark$ & $\checkmark$ & 32.81 & 24.45 & 22.77 \\
$\checkmark$ & $\checkmark$ & $\checkmark$ & $\checkmark$ & $\checkmark$ & \textbf{33.20} & \textbf{24.75} & \textbf{23.06} \\
\bottomrule
\end{tabular}
\caption{Ablation studies on COCO-C dataset with GDINO Swin-T backbone.}
\label{tab:table3}
\end{table}

\subsection{Ablation Study}
To investigate the contribution of each component in PISA, we conduct ablation studies on the COCO-C validation set with GDINO Swin-T backbone. As illustrated in Table \ref{tab:table3}, the baseline yields an mAP of 20.45\%. Introducing the main branch of FAM and $\text{BAA}_a$ improves mAP to 22.00\%. Adding the target branch of FAM yields 21.74\%, while further incorporating $\text{BAA}_a$ brings a substantial gain to 22.92\%. This suggests that while pseudo-individual source-domain features cannot directly replace source-domain features, they can be leveraged to offer effective and accurate guidance for optimizing the model at test time. Alternatively, adding $\text{BAA}_t$ to the FAM main branch with $\text{BAA}_a$ achieves 22.77\% mAP. Ultimately, the full model integrating all components achieves the best performance, reaching 33.20\% AP@50\%, 24.75\% AP@75\%, and 23.06\% mAP. This represents a cumulative improvement of 2.61\% in mAP over the baseline, demonstrating that FAM, $\text{BAA}_t$, and $\text{BAA}_a$ are mutually complementary and collectively enhance robustness against visual corruptions. We further visualize the feature distributions of clean and corrupted samples at output layer 1 using t-SNE. As illustrated in Figure \ref{tsne_ablation}, the corrupted feature distribution progressively aligns with the clean feature distribution as more modules are introduced. This indicates that the joint effect of all modules successfully transfers the features of corrupted images toward the source domain.

\subsection{Qualitative Results}
To clearly demonstrate the superiority of our proposed method, we visualize the detection results of the original GLIP and PISA under various corruptions. As shown in Figure \ref{fig:5}, when using the same backbone, compared to the original GLIP which often produces incorrect predictions or even fails to detect any objects, PISA exhibits significantly stronger robustness against different types of corruptions, yielding more accurate predictions for both object categories and bounding boxes.

\begin{figure}
\centering 
\includegraphics[width=\columnwidth]{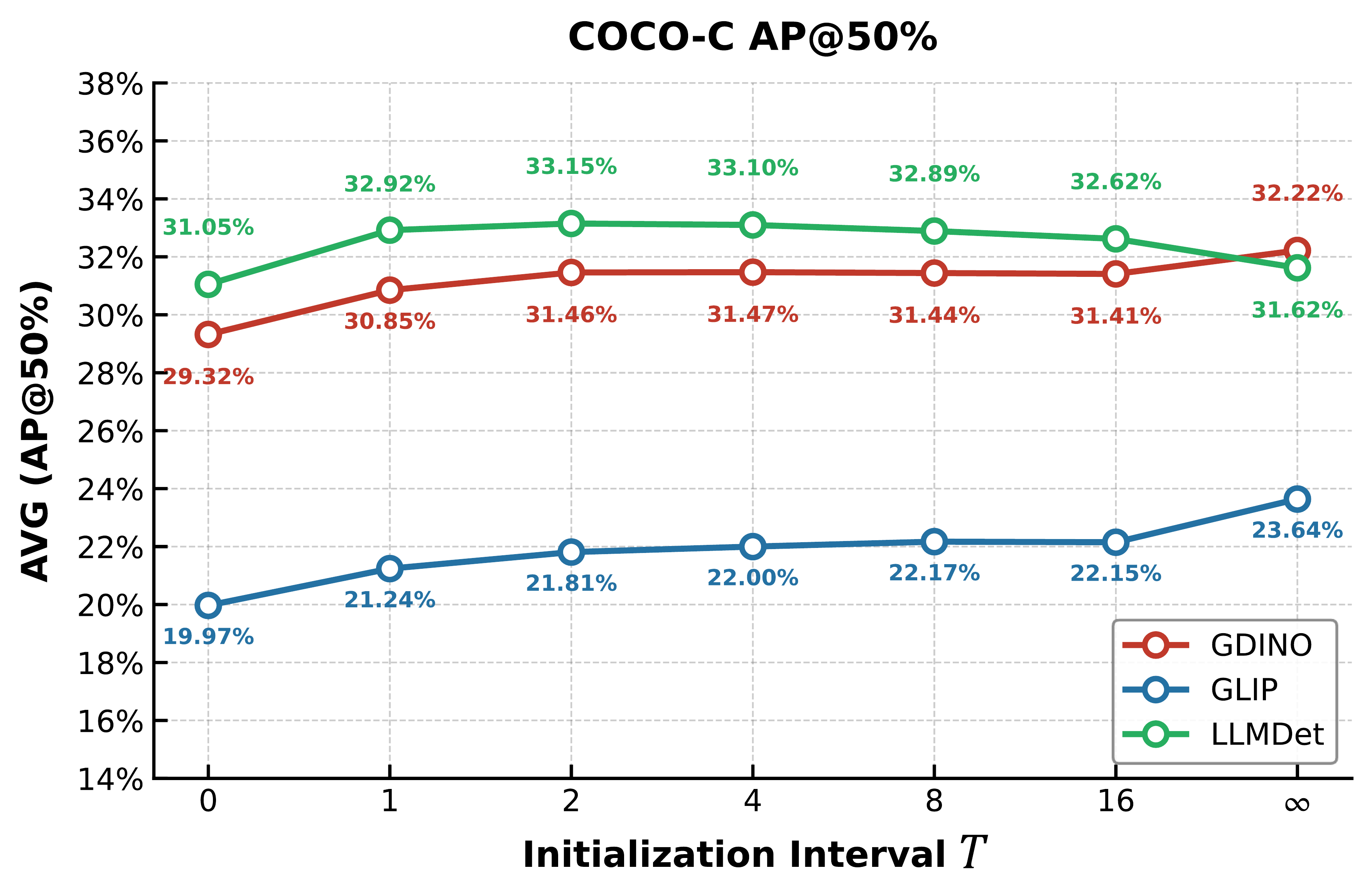} 
\caption{Average AP@50\% on COCO-C under different initialization intervals $T$ for GDINO, GLIP, and LLMDet.} 
\label{sensitivity analysis} 
\end{figure} 

\subsection{Sensitivity Analysis}
To further investigate the impact of the model initialization frequency on performance, we conduct a parameter sensitivity analysis on the initialization interval. Figure \ref{sensitivity analysis} illustrates the average metrics across 22 types of corruptions. When the initialization interval is set to 0, TTA is disabled. We evaluate initialization intervals of 1, 2, 4, 8, 16, $\infty$. It can be observed that as the initialization interval increases, the overall performance shows an upward trend and consistently outperforms the baseline without TTA. This further demonstrates that PISA, which leverages pseudo-individual source-domain features to guide model updates, is both effective and robust.

\section{Conclusion}
In this paper, we present PISA, a source-free test-time adaptation framework for open-vocabulary object detection. By leveraging corruption-invariant features from CIFE and using pseudo-individual source-domain features to adapt the model during test time, PISA reconstructs source features for individual corrupted images to effectively bridge the domain gap. This eliminates the reliance on high-quality pseudo-labels or test-time re-scoring, providing a novel solution for instance-level source-domain estimation. Furthermore, PISA is highly compatible with various backbone models, significantly improving robustness against diverse and unseen corruptions and consistently outperforming existing state-of-the-art methods on challenging benchmarks.
\appendix

\bibliography{aaai2027}


\end{document}